\documentclass{article}
\usepackage{spconf,amsmath,graphicx,hyperref}
\usepackage{booktabs}
\usepackage{multirow}
\usepackage{arydshln}

\title{Improve MLLM Benchmark Efficiency through Interview}
\address{$^{1}$Shanghai Jiao Tong University, $^{2}$Shanghai AI Laboratory  \\
wenfarong@sjtu.edu.cn}
\begin{document}
%
\maketitle

\begin{abstract}
The rapid development of Multimodal Large Language Models (MLLM) has led to a wide range of MLLM applications, and a number of benchmark datasets have sprung up in order to assess MLLM abilities. However, full-coverage Q\&A testing on large-scale data is resource-intensive and time-consuming. To address this issue, we propose the MLLM Interview (MITV) strategy, which aims to quickly obtain MLLM performance metrics by asking fewer questions. First, we constructed the interview dataset, which was built on an existing MLLM assessment dataset, by adding difficulty labels based on the performance of some typical MLLMs in this dataset. Second, we propose an MLLM Interview strategy, which obtains an initial performance situation of the large model by quizzing a small number of topics and then continuously tries to test the model's limits.
Through extensive experiments, the result shows that the MITV strategy proposed in this paper performs well on MLLM benchmark datasets, and it is able to obtain the model evaluation capability faster through a small number of questions and answers.
\end{abstract}
\section{Introduction and Related Works}

The rapid advancement of Multimodal Large Language Models (MLLMs) has significantly enhanced their ability to perform complex reasoning tasks across diverse modalities such as text, images, and beyond. Early models like CLIP-ViT \cite{radford2021learning} laid the foundation for visual-textual alignment, while more recent architectures \cite{achiam2023gpt,team2023gemini} have achieved remarkable progress in sophisticated reasoning and understanding, driven by large-scale models and extensive datasets. Notable advancements \cite{chen2024internvl,li2024llavaonevision} have further expanded the applicability of MLLMs for general-purpose and domain-transfer tasks. As MLLMs continue to evolve, the need for efficient and comprehensive evaluation methods to assess their capabilities across a wide range of tasks becomes critical.
\begin{figure}
    \centering
    \includegraphics[width=\linewidth]{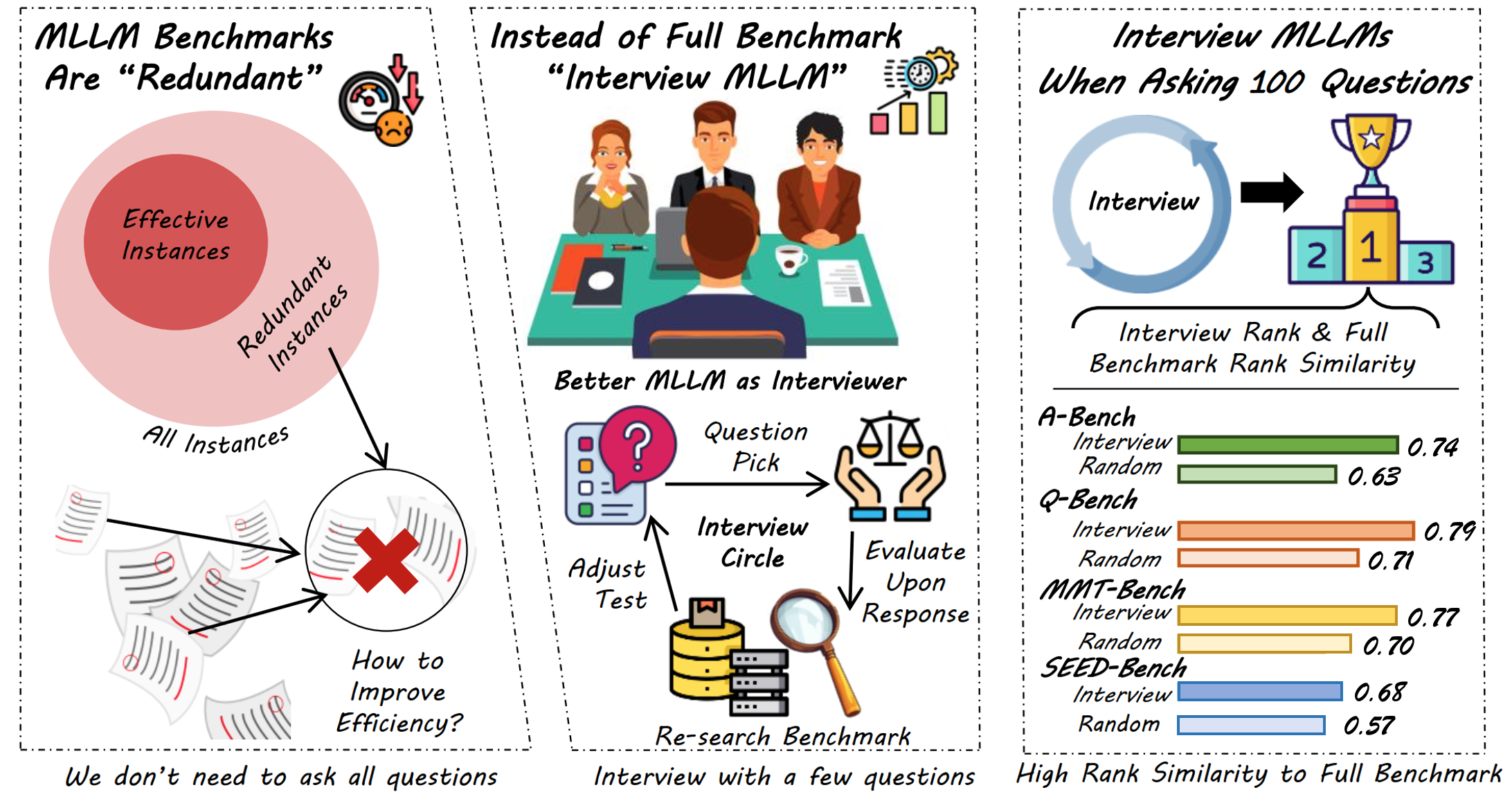}
    \caption{Motivation for our work. Inspired by human interview processes, we propose an interview strategy that dynamically adjusts questions based on MLLM performance, achieving more effective rankings than random sampling with the same number of questions.}
    \label{fig:placeholder}
\end{figure}

The evaluation of MLLMs has largely relied on benchmarks designed to assess specific capabilities like visual perception, reasoning, and domain-specific knowledge. Traditional benchmarks \cite{hudson2019gqa,antol2015vqa} typically use fixed sets of questions to evaluate models, but they often fail to capture the full complexity of generative and interactive reasoning capabilities of advanced models. In response, new benchmarks \cite{liu2025mmbench,yue2024mmmu} have been proposed to assess more integrated multimodal capabilities, while task-specific benchmarks \cite{lu2023mathvista,hu2024omnimedvqa} have focused on specialized domains.

However, the expansion of these benchmarks has introduced significant challenges, particularly redundancy and high computational costs. Models are often required to be evaluated on thousands of questions, leading to excessive time and computational expenses, while many of these questions do not sufficiently differentiate model performance. Studies indicate that evaluating MLLMs with just 40\% of benchmark instances yields rankings almost identical to those obtained using the full benchmark, suggesting that redundancy is a critical issue \cite{zhang2025redundancy}. This highlights the inefficiency of current static benchmark approaches and points to the need for more dynamic and adaptive evaluation methods.

To address these challenges, we draw inspiration from the human interview process. Experienced interviewers can assess a candidate’s abilities with a few carefully selected questions, adapting their inquiries based on the responses. This dynamic, interactive approach is more efficient than static, predefined tests. Motivated by this, we propose an interview-based evaluation framework for MLLMs, which aims to replicate the flexibility and efficiency of human interviews while maintaining the rigor of traditional benchmark assessments. 

Our approach involves the following contributions:
\textbf{(a) Construction of Interview Dataset}: We create a structured interview dataset by fusing existing benchmarks, labeling questions with difficulty and category information to form a targeted question pool.
\textbf{(b) Dynamic Interview-Based Evaluation}: A powerful MLLM acts as the interviewer, dynamically selecting questions based on the interviewee's responses. This iterative process efficiently probes the model's capabilities across various tasks and difficulty levels.
\textbf{(c) Empirical Validation}: Our experiments show that the interview-based approach (MITV) outperforms random selection strategies, providing nearly identical ranking accuracy to full benchmarks with significantly fewer questions.

This work paves the way for more efficient and practical evaluation strategies, enabling rapid assessment of MLLMs in both research and deployment contexts.

\section{Dataset Construction}
\subsection{Dataset Preparation}
\label{sec:perpare}
Traditional interview questions often lack systematic structure and difficulty gradient, which makes it difficult to comprehensively examine the interviewee's ability performance in different fields and levels. In this paper, we construct a dataset with a clear difficulty gradient to comprehensively evaluate a model's ability. This dataset fuses several existing datasets, specifically including A-Bench\cite{zhang2024a-bench}, Q-Bench\cite{wu2023q}, MMT-Bench\cite{2024mmt-icml} and SEED-Bench\cite {li2023seed}, covering multiple assessment dimensions such as logical reasoning, multimodal comprehension, multitasking, and safety ethics, which makes the data more three-dimensional.

\subsection{Difficulty Calculation}
In order to obtain the difficulty of each question quickly and fairly, the difficulty of each question was determined based on the performance of several typical MLLMs. Specifically, Duan $et$ $al$.\cite{duan2024vlmevalkit} proposed VLMEvalKit, which is an open-source evaluation toolkit of large vision-language models. VLMEvalKit provides a powerful tool that helps us test the performance of different MLLMs. For each benchmark mentioned in Section~\ref{sec:perpare}, we uniformly choose ten models, including GPT-4o\cite{openai2024gpt4o}, Deepseek-VL\cite{deepseekvl2}, Qwen-2.5-VL\cite{bai2025qwen2},Gemini-Pro-1.5\cite{team2023gemini}, Grok-3\cite{grok2025}, Kimi-VL\cite{team2025kimi}, InternVL-3\cite{chen2024internvl2},Claude-3.7-sonnet\cite{anthropic2024claude3}, Llama-3.2\cite{meta_llama3_1_2024} and Phi-3\cite{abdin2024phi3}. Then we judge the question difficulty based on the performance of the chosen MLLM according to Table~\ref{tab:Dif}, it is worth noting that questions where none of the ten models got it right are excluded.

\begin{table}[t]
\caption{The question difficulty mapping based on the number of correct responses from the MLLMs.}
\label{tab:Dif}
\centering
\resizebox{.45\textwidth}{!}{
\begin{tabular}{c|ccccc}
\toprule
Difficulty & Level1 & Level2 & Level3 & Level4 & Level5 \\
\midrule
Correct Num. &\textbf{9} & \textbf{8} & \textbf{7} & \textbf{6} & \textbf{5} \\
\midrule
Difficulty & Level6 & Level7 & Level8 & Level9 & Level10 \\
\midrule
Correct Num. & \textbf{4} & \textbf{3} & \textbf{2} & \textbf{1 }& \textbf{0}\\
\bottomrule
\end{tabular}}
\end{table}

\begin{figure}[t]
    \centering
    \includegraphics[width=\linewidth]{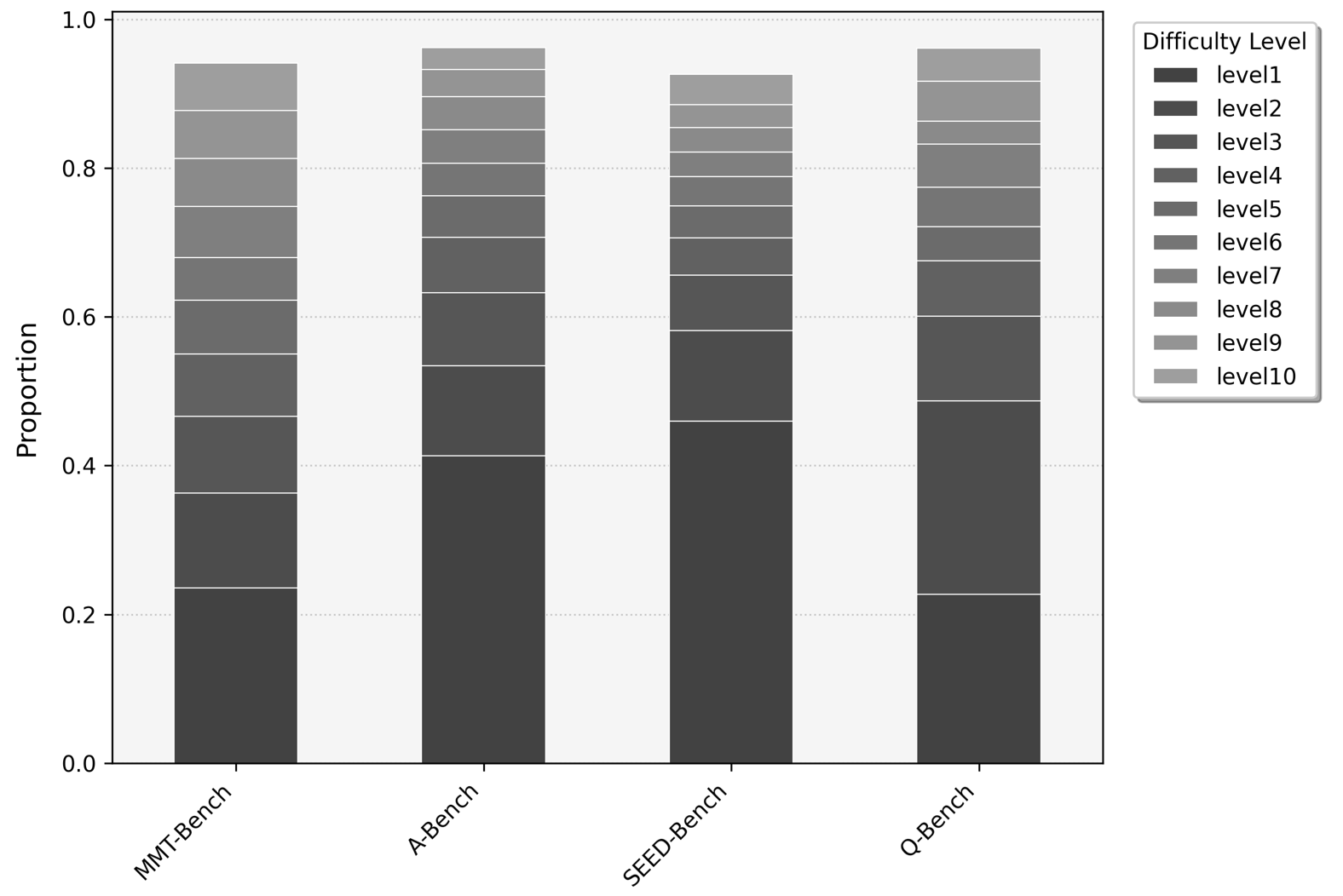}
    \caption{Difficulty distribution of questions in different benchmarks.}
    \label{fig:distru} 
\end{figure}

After processing, we analyzed the difficulty distribution across benchmarks shown in Figure~\ref{fig:distru}. The results reveal three key observations: (a) \textbf{SEED-Bench and A-Bench are overall easier}, with approximately half of the questions falling into the low-difficulty range; (b) \textbf{MMT-Bench exhibits a relatively balanced distribution} across different difficulty levels, though easier questions are still slightly more prevalent; (c) Across all four benchmarks, the number of questions that\textbf{ none of the ten models} answered correctly is small, indicating that extremely difficult items are relatively rare.

\section{Proposed Method}
\begin{figure}[t]
    \centering
    \includegraphics[width=\linewidth]{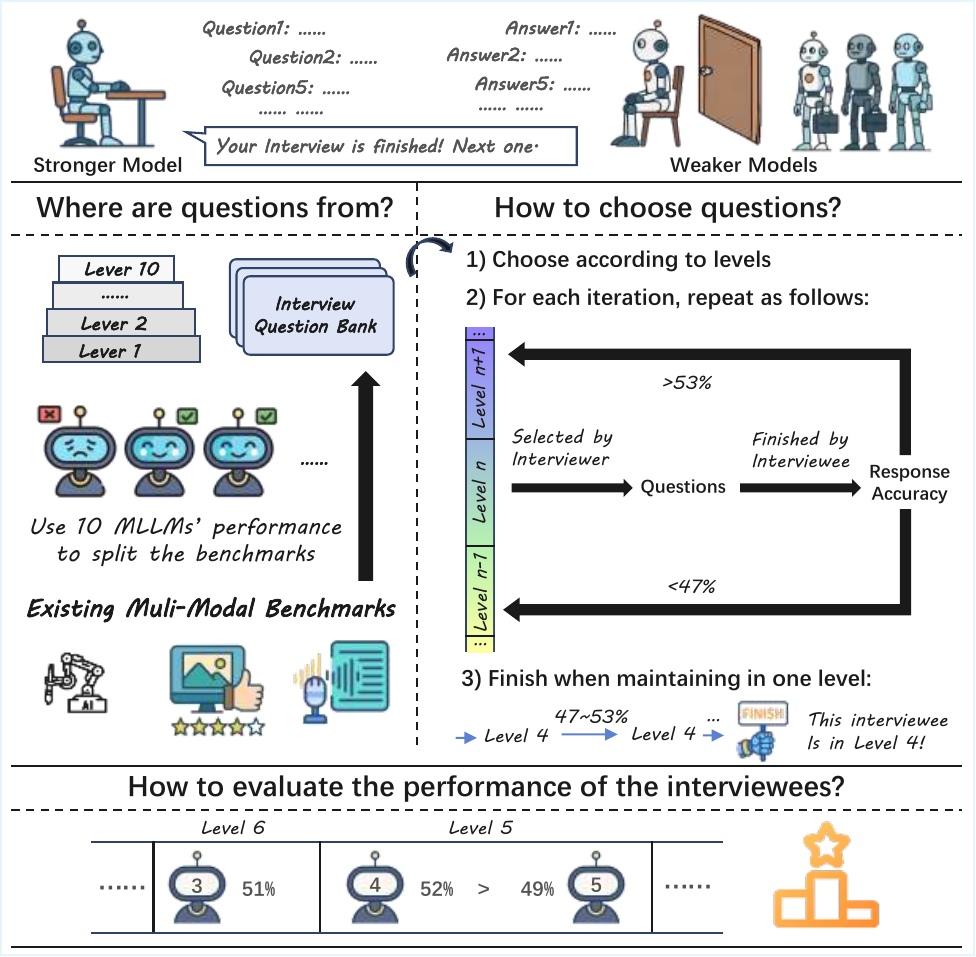}
    \caption{The proposed MITV strategy framework.}
    \label{fig:frame}
    \vspace{-0.2cm}
\end{figure}

In this section, we introduce the proposed MITV strategy, whose framework is illustrated in Figure \ref{fig:frame}. It comprises three main modules: the question selection module, the interview module, and the result evaluation module. First, the question selection module determines the difficulty level of the next question based on the respondent's answers. Next, the interview module presents the selected question to the respondent for a response. Finally, the result analysis module evaluates the respondent's answers, analyzes the correctness of the responses, and ultimately summarizes the respondent's performance with a final score as the final outcome.

\subsection{Difficulty Determination}
The difficulty determined module integrates information theory and adaptive testing theory. Its core objective is to dynamically adjust question difficulty to efficiently maximize information gain about the model's capabilities.
Let the outcome of a question at difficulty level $l$ be $Y\in\{0,1\}$ (correct/incorrect) with success probability ${p}_l$. The expected information (Bernoulli entropy):
\begin{equation}
   H(p_l)=-\,p_l\log p_l-(1-p_l)\log(1-p_l) 
\end{equation}
where $H(p_l)$ is maximized at $p_l=0.5$. Hence, the level at which the model attains $\approx 50\%$ accuracy is \emph{ability-aligned}: items there are most informative about the model’s competence. In our design, we \emph{do not} compute information for each question; instead, this principle motivates a simple controller that steers the process toward the $p_l\!\approx\!0.5$ regime.

To rapidly localize the interviewee model’s ability, we initialize the interview at the mid difficulty (\(l=5\)). After each response, we update the accuracy at the current level and adjust the difficulty of the next item accordingly; the procedure is given by:
\begin{equation}
l_{t+1}=
\begin{cases}
\min(l_t+1,\,L_{\max}), & \text{if } {p}_{l_t}>0.52,\\[2pt]
\max(l_t-1,\,L_{\min}),        & \text{if } {p}_{l_t}<0.48,\\[2pt]
l_t,                    & \text{others.}
\end{cases}
\end{equation}
where $l_t$ indicates the current difficulty level, $l_{t+1}$ indicates the next difficulty level, $L_{\max}$ and $L_{min}$ denote the maximum and minimum difficulty levels, ${p}_{l_t}$ represents the accuracy rate of interviewee answers in level $l$.

\begin{table}[t]\small
    \centering
    \renewcommand{\arraystretch}{1.1}
    \caption{MLLM Interviewees Illustration. We utilize 9 closed-source MLLM interviewees accessed via API calls and 10 open-source MLLM interviewees deployed locally.}
    \begin{tabular}{c|l}
        \toprule
        \textbf{Calling Method} & \textbf{Model} \\
        \hline 
        \multirow{3}{*}{API Call} 
            &Gpt-4.1-Nano \cite{achiam2023gpt}, Gpt-4o-Mini \cite{openai2024gpt4o} \\
            &Gpt-4.1 \cite{achiam2023gpt}, Gpt-4o \cite{openai2024gpt4o}, Grok-3 \cite{grok2025}\\
            &Claude-3.7-Sonnet \cite{anthropic2024claude3}, Claude-3.5-Sonnet \cite{anthropic2024claude3}\\
            &Qwen-VL-max \cite{bai_qwen_vl_2023}, Qwen-VL-plus \cite{bai_qwen_vl_2023}\\
        \hline
        \multirow{5}{*}{Local Call} 
            & Phi-3.5 \cite{abdin2024phi3}, Phi-3 \cite{abdin2024phi3},Qwen2.5-VL-7b\cite{bai2025qwen2} \\
            & Qwen2.5-VL-72b \cite{bai2025qwen2}, Qwen2.5-VL-32b \cite{bai2025qwen2}\\
            & InternVL2-4b \cite{chen2024internvl2}, InternVL2.5-4b \cite{chen2024internvl2} \\
            & InternVL3-8b \cite{chen2024internvl2}, Mini-InternVL \cite{openxlab_mini_internvl_2024} \\
            & Llama-3.2-11b-Vision-Instruct \cite{meta_llama3_1_2024}\\
        \bottomrule
    \end{tabular}
    \label{tab:selectmodel}
    \vspace{-0.2cm}
\end{table}

\begin{table*}[t] 
  \centering
  \renewcommand\arraystretch{1.15}
  \renewcommand\tabcolsep{6pt}
  \caption{Performance comparison between the random and the proposed interview strategy, where `Question Num.' indicates the number of used questions. }
    \resizebox{\linewidth}{!}{\begin{tabular}{c|rrr|rrr|rrr|rrr}
    \hline
    \hline
     \textbf{Benchmark}  & \multicolumn{3}{c|}{A-Bench} &  \multicolumn{3}{c|}{Q-Bench} & \multicolumn{3}{c}{MMT-Bench} & \multicolumn{3}{|c}{SEED-Bench}\\ \hline
     \textbf{Question Num.}  & SRCC & PLCC & KRCC & SRCC & PLCC & KRCC & SRCC & PLCC & KRCC & SRCC & PLCC & KRCC \\
    \hline
    \multicolumn{13}{l}{\textit{Random Strategy: Questions are randomly sampled from the benchmarks.}} \\ \hdashline
     10  &  0.3058  &  0.3316  &  0.1846  &  0.2767  &  0.3356  &  0.1657  &  0.3330  &  0.4717  &  0.2094  &  0.2677  &  0.5912  &  0.1633\\
    20  &  0.3841  &  0.4287  &  0.2480  &  0.4773  &  0.4947  &  0.3073  &  0.4239  &  0.6103  &  0.2887  &  0.2933  &  0.6939  &  0.1756\\
    30  &  0.4432  &  0.6000  &  0.2916  &  0.4794  &  0.5834  &  0.3207  &  0.5460  &  0.6844  &  0.3800  &  0.4265  &  0.7098  &  0.2772\\
    50  &  0.5021  &  0.6303  &  0.3406  &  0.5591  &  0.6498  &  0.3909  &  0.6246  &  0.7355  &  0.4587  &  0.4502  &  0.7566  &  0.3095\\
    100  &  0.6365  &  0.7375  &  0.4589  &  0.7114  &  0.7388  &  0.5308  &  0.7046  &  0.7730  &  0.5358  &  0.5776  &  0.7897  &  0.4057\\
    \hline
    \multicolumn{13}{l}{\textit{Interview Strategy (proposed): Questions are picked during the interview process.}} \\
    \hdashline
10  &  0.3958  &  0.4458  &  0.2858  &  0.3667  &  0.4356  &  0.2597  &  0.4537  &  0.5417  &  0.3294  &  0.3577  &  0.6412  &  0.2533\\
20  &  0.4741  &  0.5287  &  0.3480  &  0.5673  &  0.5847  &  0.3973  &  0.5139  &  0.6603  &  0.3787  &  0.3833  &  0.7439  &  0.2656\\
30  &  0.5532  &  0.6700  &  0.4016  &  0.5894  &  0.6734  &  0.4207  &  0.6160  &  0.7344  &  0.4700  &  0.4965  &  0.7598  &  0.3672\\
50  &  0.6121  &  0.7103  &  0.4506  &  0.6491  &  0.7198  &  0.4809  &  0.6946  &  0.7855  &  0.5487  &  0.5702  &  0.8066  &  0.3995\\
100  &  0.7465  &  0.8175  &  0.5689  &  0.7914  &  0.8288  &  0.6208  &  0.7746  &  0.8230  &  0.6258  &  0.6876  &  0.8397  &  0.4957\\
    \hline 
    \hline
    \end{tabular}}
  \label{tab:addlabel}
\end{table*}

\subsection{MLLM Interview Module}
In order to make the interview questions more effective and representative, the module selects 10 different types of questions based on the target level. The interviewer MLLM selects a representative question from ten based on the test previous responses, then passes this question to the candidate MLLM for answering. Finally, the interviewer evaluates the candidate's answers for correctness. Through this systematic process of question screening and interaction, it not only ensures that the interview questions can accurately match the assessment needs, but also flexibly adjusts the direction of the investigation based on the test MLLM feedback, which helps to assess its competency level in a more comprehensive manner and improves the accuracy of the interview assessment.


\subsection{Data Process Module}
\label{subsec:judgment}
To evaluate the interviewee model, we define its capability level $l_*$ as the difficulty level where the model achieves an accuracy close to $50\%$ within a small tolerance $\epsilon$:

\begin{equation}
l_* = \max \Big\{\, l \;\big|\; {p}_l \in [0.5 - \epsilon,\; 0.5 + \epsilon] \,\Big\},
\end{equation}
where ${p}_l$ denotes the accuracy of the model at difficulty level $l$. Based on this capability level, the final performance score $S$ of the tested MLLM is computed as:

\begin{equation}
\!S\! =\!
\begin{cases}
0, & p_{l_*} \!<\! 0.5\! -\! \epsilon \;\text{and}\; l_* = 0, \\[6pt]
0.1\,l_* \!+\! \frac{0.1}{\epsilon}\big({p}_{l_*}\! - \!(0.5 \!-\! \epsilon)\big), & {p}_{l_*} \in [0.5 - \epsilon,0.5 + \epsilon], \\[6pt]
1, & {p}_{l_*}\! >\! 0.5 \!+ \!\epsilon \;\text{and}\; l_* = 9.
\end{cases}
\end{equation}
where $l_*$ reflects the model’s capability-aligned difficulty, and $S$ denotes the normalized final performance score.


\section{Experiment}
\subsection{Experiment Detail}
In order to validate the generalization of MITV, we have selected 19 typical MLLMs, as detailed in Table \ref{tab:selectmodel}. Among them, 9 large models were validated using official API calls, and for the other MLLM models, we used locally deployed models for the validation. To verify the validity of MITV, we designed a control group whose Benchmark questions were quizzed through a random sampling strategy. It is worth noting that the random sampling strategy group experimental is tested by VLMEvalKit \cite{duan2024vlmevalkit}. Due to the balanced performance of Gpt-4o~\cite{openai2024gpt4o} across various tasks, we adopt Gpt-4o as the interviewer model. we set the tolerance parameter $\epsilon = 0.02$.

On each Benchmark, the full coverage test performance of the model is used as the ground truth and three commonly used metrics for algorithm assessment are applied: Spearman rank order correlation coefficient (SRCC), Pearson linear correlation coefficient (PLCC), and Kendall rank order correlation coefficient (KRCC). 
\vspace{-0.3cm}
\subsection{Performance Discussion}
The experimental results in Table \ref{tab:addlabel} demonstrate the effectiveness of the proposed MITV strategy compared to the random sampling baseline across four MLLM benchmarks. With closer inspection, we can obtain several insights as follows:

\textbf{Superior Performance of MITV:}
Across all benchmarks, MITV achieves higher SRCC, PLCC, and KRCC scores than the random sampling strategy in most settings. For instance, on A-Bench with 100 questions, MITV yields an SRCC of 0.7465, compared to 0.6365 for random sampling with a 17.4\% improvement. These results highlight MITV’s ability to produce rankings more aligned with full benchmark evaluations, even with a limited number of questions.

\textbf{Efficiency with Small Question Sets:}
When the number of questions is small, MITV demonstrates a significant advantage. For example, in the MMT-Bench test, MITV achieved an SRCC of 0.4537 with 10 questions, while random sampling yielded only 0.3330 with a 36.0\% improvement. This highlights the efficiency of MITV's adaptive question selection mechanism: by leveraging difficulty metadata and performance feedback, it can precisely target information-rich questions early in the evaluation process. In contrast, random sampling struggles to capture meaningful performance differences when the number of questions is limited.

\textbf{Generalization Across Benchmarks:}
The performance gains of MITV are consistent across diverse benchmarks, demonstrating its generalizability. On SEED-Bench, which has a larger benchmark, MITV achieves an SRCC of 0.6876 with 100 questions, compared to 0.5776 for random sampling. On smaller benchmarks like Q-Bench and MMT-Bench, MITV maintains its superiority, with SRCC values exceeding 0.77 at 50 questions.

\section{Conclusion}
Since the conventional Benchmark test is a full-coverage question and answer test, there is information redundancy, in order to optimise the evaluation method, this paper firstly constructs several datasets with difficulty labels through the performance of ten models on the existing Benchmark.  Then the MITV strategy is proposed, which can obtain the fastest model evaluation performance through a small number of questions and answers by converting the conventional full-coverage model performance test into an interview ability evaluation.  Experiments prove that the proposed method is effective, has good generalisation, and can provide suggestions and guidance for MLLM assessment work.

\bibliographystyle{IEEEbib}
\bibliography{strings,refs}

\end{document}